\documentclass[conference]{IEEEtran}
\IEEEoverridecommandlockouts
\usepackage{cite}
\usepackage{amsmath,amssymb,amsfonts}
\usepackage{algorithmic}
\usepackage{graphicx}
\usepackage{textcomp}
\usepackage{times}
\usepackage{latexsym}
\usepackage{multirow}
\usepackage{siunitx}
\usepackage{pifont}
\newcommand{\cmark}{\ding{51}}%
\newcommand{\xmark}{\ding{55}}%
\usepackage[table,xcdraw]{xcolor}
\usepackage{colortbl}
\usepackage{placeins}
\usepackage{mathtools}
\usepackage{xcolor}

\def\BibTeX{{\rm B\kern-.05em{\sc i\kern-.025em b}\kern-.08em
    T\kern-.1667em\lower.7ex\hbox{E}\kern-.125emX}}
\newif\ifmark
\newcommand{\changemarker}[1]{%
\ifmark
\textcolor{blue}{#1}%
\else
#1%
\fi
}

\makeatletter
\def\ps@IEEEtitlepagestyle{
  \def\@oddfoot{\mycopyrightnotice}
  \def\@evenfoot{}
}
\def\mycopyrightnotice{
  {\footnotesize
  \begin{minipage}{\textwidth}
  ~\copyright~2020 IEEE. Personal use of this material is permitted. Permission from IEEE must be obtained for all other uses, in any current or future media, including reprinting/republishing this material for advertising or promotional purposes, creating new
collective works, for resale or redistribution to servers or lists, or reuse of any copyrighted component of this work in other works. DOI: 10.1109/IJCNN48605.2020.9207209
  \end{minipage}
  }
}

\begin{document}
\markfalse 
\title{Simple and Effective Prevention of Mode Collapse in Deep One-Class
Classification
\thanks{} 
}


\author{\IEEEauthorblockN{Penny Chong\IEEEauthorrefmark{1}
,
Lukas Ruff\IEEEauthorrefmark{2},
Marius Kloft\IEEEauthorrefmark{4}, and Alexander Binder\IEEEauthorrefmark{1}
}\\
\IEEEauthorblockA{
\IEEEauthorrefmark{1}ISTD Pillar, Singapore University of Technology and Design, Singapore\\
\IEEEauthorrefmark{2}Machine Learning Group, Technische Universität Berlin, Berlin, Germany\\
\IEEEauthorrefmark{4}Department of Computer Science, Technische Universität Kaiserslautern, Kaiserslautern, Germany\\
Email: penny\_chong@mymail.sutd.edu.sg
}}
\maketitle

\begin{abstract}
Anomaly detection algorithms find extensive use in various fields. This area of research has recently made great advances thanks to deep learning. A recent method, the deep Support Vector Data Description (deep SVDD), which is inspired by the classic kernel-based Support Vector Data Description (SVDD), is capable of simultaneously learning a feature representation of the data and a data-enclosing hypersphere. The method has shown promising results in both unsupervised and semi-supervised settings. However, deep SVDD suffers from hypersphere collapse---also known as mode collapse---, if the architecture of the model does not comply with certain architectural constraints, e.g. the removal of bias terms. These constraints limit the adaptability of the model and in some cases, may affect the model performance due to learning sub-optimal features. In this work, we consider two regularizers to prevent hypersphere collapse in deep SVDD. 
The first regularizer is based on injecting random noise via the standard cross-entropy loss. The second regularizer penalizes the minibatch variance when it becomes too small. Moreover, we introduce an adaptive weighting scheme to control the amount of penalization between the SVDD loss and the respective regularizer. Our proposed regularized variants of deep SVDD show encouraging results and outperform a prominent state-of-the-art method on a setup where the anomalies have no apparent geometrical structure.
\end{abstract}


\section{Introduction}
Anomaly detection \cite{chandola2009anomaly} has long been an active area of research \cite{ramadas2003detecting,zhai2016deep}. It refers to the detection of unusual events or instances in the data that do not conform to the normal behavior \cite{chandola2009anomaly}. 
Anomaly detection algorithms are used extensively in a wide range of applications such as network intrusion detection systems \cite{shone2018deep}, fraud detection systems \cite{huang2018codetect}, and the detection of cyber attacks in Cyber-Physical systems (CPS) \cite{goh2017anomaly}. In principal, one can frame anomaly detection as a binary supervised classification problem, but such an approach has several disadvantages. An algorithm which learns to classify directly the normal versus anomalous samples has no guarantee to generalize well to new types of anomalies or threats. It is also challenging to collect all possible types of anomalies/threats due to their diversity, lack of labels, or the cost to obtain. Addressing this, Steinwart et al. (2005) \cite{steinwart2005classification} assume a uniform distribution for anomalous points, but their approach suffers from the curse of dimensionality. To detect new anomalies, many of these systems adopt an \emph{unsupervised} learning approach as foundation for their detection algorithm.
Some approaches involve detection based on reconstruction errors \cite{Zhou:2017:ADR:3097983.3098052,schlegl2017unsupervised} and some are based on principles derived from classification approaches \cite{scholkopf2000support, tax2004support,liu2008isolation}. 

Recently \cite{ruff2018deep} proposed an unsupervised one-class classification approach---deep SVDD, which jointly learns the feature representation of the data and a data-enclosing hypersphere. The unsupervised deep SVDD model has been generalized to a  semi-supervised deep SVDD model \cite{ruff2019deep}. 
However, both models suffer from \emph{hypersphere collapse} if the architecture of the model does not comply with a set of architectural constraints. The imposed set of constraints include the use of only unbounded activation functions and the removal of bias terms in the network architecture \cite{ruff2018deep}. Imposing these restrictions may cause the network to learn sub-optimal features only. 
A conceptually related problem to hypersphere collapse is the mode collapse problem observed in GANs \cite{DBLP_conf_iclr_ArjovskyB17,salimans2016improved}. In GANs, the generated samples tend to collapse to a limited number of modes without proper regularization. In the case of deep SVDD, the learned features collapse to a single point if the architectural constraints are not enforced.
We use the two terms---mode and hypersphere collapse---interchangeably in this work. We provide related work on the mode collapse problem in GANs in Section \ref{RELATED WORKS}.

In this paper, we address the hypersphere collapse problem in deep SVDD as follows. We propose two regularizers based on random noise injections and minibatch variance, respectively. The resulting regularized variants of deep SVDD form a simple, yet effective regularization against hypersphere collapse in deep SVDD, thereby allowing the removal of architectural constraint on the bias term and activation.
The following summarizes our contributions in this work.
\begin{enumerate}
\item We propose a regularization technique against hypersphere collapse based on noise injection via the standard cross-entropy loss.
\item We propose a regularization technique against hypersphere collapse that penalizes the minibatch variance if it falls below some annealed threshold. 
\item We propose an adaptive weighting scheme to control the balance between the deep SVDD loss and the regularizer.
\item We show that our regularized deep SVDD variants generally perform better than the standard deep SVDD and other methods for both unsupervised and semi-supervised settings. 
\end{enumerate}

\section{Related Work}\label{RELATED WORKS}
\noindent\textbf{Anomaly detection.}
 \ Approaches in this area can be broadly divided into reconstructive and classification-based methods. 
 For reconstructive methods, usually with autoencoders or GANs, anomalies can be detected from analysing the reconstruction errors since the model only learns to reconstruct normal samples during training.
 In \cite{Zhou:2017:ADR:3097983.3098052}, the authors proposed the robust deep autoencoders (RDA) method which is a combination of autoencoders and robust principle component analysis (RPCA) \cite{candes2011robust} for anomaly detection. 
 A more advanced approach \cite{beggel2019robust} handles the dataset contamination problem by adapting an adversarial autoencoder architecture to impose a prior distribution on the latent representation by placing anomalies into low likelihood-regions. On the other hand, the authors in \cite{schlegl2017unsupervised} are the first to employ a GAN approach for anomaly detection. They introduce the unsupervised AnoGAN model to learn a manifold of the normal samples with generative and discriminative components.
Other works that adopt GANs for anomaly detection are \cite{deecke2018image,zenati2018efficient,akcay2018ganomaly}.

Another class of anomaly detection algorithms are classification-based approaches. 
The isolation forest (IF) \cite{liu2008isolation} and the one-class support vector machine (OC-SVM) \cite{scholkopf2000support} are unsupervised methods. The kernel-based SVDD \cite{tax2004support} which learns a data-enclosing hypersphere of a minimum volume in feature space is inspired by the OC-SVM. The deep SVDD model \cite{ruff2018deep} improves the kernel-based SVDD model by simultaneously learning the feature representation of the unlabeled data and the minimum volume data-enclosing hypersphere. The unsupervised deep SVDD method has also been generalized to the semi-supervised setting \cite{ruff2019deep}. In \cite{golan2018deep}, the authors learn to solve the auxiliary task of classifying different geometric transformations of the normal data. The anomalies are then identified based on their softmax activation statistics of the transformed images. Their approach achieves a new state-of-the-art performance on the CIFAR10 dataset, significantly outperforming some prominent reconstruction and classification-based approaches, including the standard deep SVDD model \cite{ruff2018deep}.
\\\\
\textbf{Mode collapse in GANs.}
\ The hypersphere collapse problem in deep SVDD is conceptually similar to the mode collapse problem in GANs. Deep generative models such as GANs, are known to be difficult to train, unstable, and prone to mode collapse \cite{DBLP_conf_iclr_ArjovskyB17,salimans2016improved}. Without proper regularization, the generator often collapses and thus, generates only samples of a few modes or a single mode of class. In \cite{che2016mode}, the authors argue that the training issues in GANs are due to the lack of control over the discriminator, and the naturally disjoint manifolds of the actual and generated data. They add a mode regularizer to the generator's optimization target in an attempt to encourage the generated samples to move towards a nearby mode of the data generating distribution. On the other hand, \cite{salimans2016improved} introduces a set of training methods such as the minibatch discrimination method to improve the training of GANs. 
Subsequently, \cite{srivastava2017veegan} proposes an additional reconstructor network that learns to map from the input data distribution to Gaussian random noise. By training the generator and reconstructor networks simultaneously via an implicit variational principle in VEEGAN, the reconstructor network learns the mapping from the data distribution to Gaussian and an approximate inverse of the generator, thereby preventing mode collapse. The work \cite{chang2018escaping} proposes a latent-space constraint loss to prevent mode collapse in Boundary Equilibrium Generative Adversarial Network (BEGAN) \cite{berthelot2017began}. 

In this work, we compare our proposed regularized variants of deep SVDD with the state-of-the-art method \cite{golan2018deep} as well as other classification-based anomaly detection methods.

\section{Preliminaries}\label{PRELIMINARIES}
Deep SVDD \cite{ruff2018deep,ruff2019deep} is an extension of the kernel-based SVDD method \cite{tax2004support}. The kernel-based SVDD learns a data-enclosing hypersphere of minimum volume in feature space from the unlabeled training data. However, the kernel method requires manual feature engineering and scales poorly due to the computation/manipulation of the kernel matrix. Deep SVDD overcomes these limitations by jointly learning a suitable feature representation of the data and mapping them to a minimum volume hypersphere \cite{ruff2018deep,ruff2019deep}. Nonetheless, under certain conditions, deep SVDD is not fully robust against mode collapse, which will be elaborated in Section \ref{RESULTS}.

\subsection{Unsupervised Deep SVDD}
For some input space $\mathcal{X} \subseteq \mathbb{R}^d$ and output space $\mathcal{F} \subseteq \mathbb{R}^p$, let $\phi(\cdot\ ;\mathcal{W}):\mathcal{X} \rightarrow \mathcal{F}$ be a neural network with $L$ hidden layers and a set of weights $\mathcal{W} = \{W^1, ..., W^L\}$ where $W^l$ are the weights of layer $l \in \{1, ..., L\}$. Therefore, $\phi(x ;\mathcal{W}) \in \mathcal{F}$ is the feature representation of $x \in \mathcal{X}$. Unsupervised deep SVDD jointly learns the feature representation and minimizes the volume of a hypersphere in $\mathcal{F}$. Given the unlabeled training data $\mathcal{D}_n = \{ x_1, ..., x_n\}$, the \textit{soft-boundary} objective function is given as
\begin{multline}
\min_{\mathcal{R,W}} R^2 + \frac{1}{\nu n}\sum^{n}_{i=1} \max\{0,||\phi(x_i ;\mathcal{W})-c ||^2 - R^2 \}\\+ \frac{\lambda}{2} \sum^{L}_{l=1} || W^l||^{2}_F \ \  
\label{unsupervised_softboundary}
\end{multline}
where $R$ is the radius, $c \in \mathcal{F}$ is the center of the hypersphere, and $\nu \in (0,1]$ is the trade-off between the volume and boundary violations of the hypersphere, i.e. fraction of outliers. $c$ is fixed as the mean of the mapped data from an initial forward pass \cite{ruff2018deep}. For every iteration, $R$ is computed as the $(1-\nu)$-th quantile of $||\phi(x_i ;\mathcal{W})-c ||_2 $.  
Assuming most of the training data is normal, objective \eqref{unsupervised_softboundary} can be simplified to the following \textit{one-class} objective function
\begin{equation}
\min_{\mathcal{W}}  \frac{1}{n}\sum^{n}_{i=1} ||\phi(x_i ;\mathcal{W})-c ||^2 +   \frac{\lambda}{2} \sum^{L}_{l=1} || W^l||^{2}_F.
\label{unsupervised_one-class}
\end{equation}

\subsection{Semi-supervised Deep SVDD}
In the semi-supervised approach, the model has access to both unlabeled training samples $ x_1, ..., x_n \in \mathcal{X}$, and labeled training samples $(\tilde{x}_1,\tilde{y}_1), ...,(\tilde{x}_m,\tilde{y}_m) \in \mathcal{X}\times\mathcal{Y}$, where $\mathcal{X}\subseteq\mathbb{R}^d$ and $\mathcal{Y}=\{-1, +1\}$. Normal samples are labeled as $\tilde{y}=+1$ and anomalies as $\tilde{y}=-1$. The \textit{one-class} semi-supervised Deep SVDD objective function is formulated as
\begin{multline}
\min_{\mathcal{W}}\frac{1}{n+m}\sum^{n}_{i=1} ||\phi(x_i ;\mathcal{W})-c||^2\\ +\frac{\eta}{n+m}\sum^{m}_{j=1}(||\phi(\tilde{x}_j ;\mathcal{W})-c ||^2)^{\tilde{y}_j}.
\label{semi-supervised_one-class}
\end{multline}
We omitted the weight decay regularizer in the above formulation for simplicity. For more information, we refer the reader to the work \cite{ruff2018deep,ruff2019deep}.

\section{Proposed Methodology}
\label{methodology}
The hypersphere collapse phenomenon and the architectural constraints for deep SVDD are briefly described in Section \ref{Manifold collapse}. The two different regularization methods are presented in Section \ref{Reg with noise inject aux} and Section \ref{Reg with minibatch var}, respectively. Section \ref{Adaptive weighting scheme} discusses the adaptive weighting scheme used to control the scaling between the deep SVDD loss and the respective regularizer.

\subsection{Hypersphere Collapse}\label{Manifold collapse}
According to \cite{ruff2018deep,ruff2019deep}, an improperly formulated deep SVDD network (which includes the hypersphere center $c$), may lead to the learning of a trivial solution, i.e. hypersphere/mode collapse, a phenomenon where the features learned by the network converge to a single point, $\phi \equiv c $. It has been recommended that the center $c$ must be something other than the all-zero-weights solution, and the network should use only unbounded activations and omit bias terms to avoid a hypersphere collapse. However, omitting bias terms in a network may only lead to a sub-optimal feature representation due to the role of bias in shifting activation values.
In this work, we propose two simple regularization techniques against mode collapse in deep SVDD which allow the revoke of previous restrictions on the class of networks.

\subsection{Regularization with Random Noise Injection via the Loss Layer} \label{Reg with noise inject aux}
We revisit the timeless idea of noise injection in machine learning, 
and, inspired by 
methods which address the overfitting issue in convolutional neural networks (CNNs) \cite{chen2017noisy,xie2016disturblabel}, we took a similar regularization approach for a loss layer 
to prevent mode collapse in deep SVDD. We added a new loss layer on top of the feature extractor to learn $k$ randomly labeled classification tasks in conjunction with the SVDD objective function.
For some input space $\mathcal{X} \subseteq \mathbb{R}^d$, feature space $\mathcal{F} \subseteq \mathbb{R}^p$, and output space $\mathcal{K} \subseteq \mathbb{R}^k$, let $\phi(\cdot\ ;\mathcal{W}_0):\mathcal{X} \rightarrow \mathcal{F}$ be a neural network with $L$ hidden layers, and $g(\cdot\ ;\mathcal{W}_1):\mathcal{F} \rightarrow \mathcal{K}$ be the linear layer after it. The final network is given by $z(x;\mathcal{W}) = g(\phi(x;\mathcal{W}_0);\mathcal{W}_1) \in \mathcal{K} $, where $\phi(x ;\mathcal{W}_0) \in \mathcal{F}$ is the feature representation of $x \in \mathcal{X}$. For a given input sample $x_i$, we define $z_{i,q}$ as the $q$-th element of the $k$-dimension logits vector $z(x_i;\mathcal{W})$ and $\sigma$ as the sigmoid function. 
We add the following sigmoid cross-entropy loss as a regularizer to the standard unsupervised or semi-supervised deep SVDD loss $\mathcal{L}_{SVDD}$ (Eq. \eqref{unsupervised_softboundary}, \eqref{unsupervised_one-class} or \eqref{semi-supervised_one-class}).
\begin{multline}
\mathcal{L}_{reg}= -\frac{1}{n k}\sum^{n}_{i=1}\sum^k_{q=1} y_{rnd} \log \sigma(z_{i,q})\\+ (1-y_{rnd})\log(1-\sigma(z_{i,q})).
\label{k-aux}
\end{multline}
This approach is based on the idea that learning a single classification problem with meaningful, non-random labels would prevent mode collapse orthogonal to the decision boundary.
Therefore, when given random binary labels, the classifier aims at preventing mode collapse along a randomly varying set of hyperplanes, 
which may potentially lead to better convergence. 
Here, at each iteration, the binary labels $y_{rnd}$ are randomly sampled from the uniform distribution of $\{0, 1\}$ for each sample. 
At test time, the linear layer $g(\cdot\ ;\mathcal{W}_1)$ will be removed. We present the overall loss function used in the training of the regularized deep SVDD in Section \ref{Adaptive weighting scheme}.

\subsection{Regularization with Hinge Loss on Minibatch Variance}\label{Reg with minibatch var}
We also consider using the hinge loss \cite{rosasco2004loss} to penalize the minibatch variance if it falls below a specified threshold. This ensures the features in a minibatch to differ sufficiently, thereby preventing mode collapse. Being able to specify the threshold also gives us more control over the regularization as compared to the method presented in Section \ref{Reg with noise inject aux} which is based on random noise injection.
Using the notations from Section \ref{Reg with noise inject aux}, we define $\phi_{i,q}$ as the $q$-th feature from the feature vector $\phi(x_i ;\mathcal{W}_0) \in \mathcal{F}$ of the input sample $x_i$. We do not add a linear layer after the feature extractor $\phi(\cdot\ ;\mathcal{W}_0)$ as done in Section \ref{Reg with noise inject aux}. We add the following regularization loss to the standard unsupervised or semi-supervised deep SVDD loss $\mathcal{L}_{SVDD}$ (Eq. \eqref{unsupervised_softboundary}, \eqref{unsupervised_one-class} or \eqref{semi-supervised_one-class}). 


\begin{align}
\mathcal{L}_{reg} &= \max\{0, d_{\tilde{t}} - \frac{1}{p(n-1)}\sum_{q=1}^{p}\sum_{i=1}^{n}(\phi_{i,q}-\overline{\phi}_q)^2 \} \\ \nonumber
&= \max\{0, d_{\tilde{t}} - \frac{1}{p(n-1)}\sum_{q=1}^{p}\sum_{i=1}^{n}(\phi_{i,q}-\frac{1}{n}\sum_{k=1}^{n}\phi_{k,q})^2 \}.
\label{loss_var_reg}
\end{align}
The term $\frac{1}{p(n-1)}\sum\limits_{q=1}^{p}\sum\limits_{i=1}^{n}(\phi_{i,q}-\overline{\phi}_q)^2$ computes the sample variance of the minibatch of features averaged over $p$ dimensions (the dimension of the feature vector) and
$d_{\tilde{t}}$ refers to the variance threshold used for penalization at epoch $\tilde{t}$. This formulation enforces a penalty only when the minibatch variance falls below the threshold, i.e. $\frac{1}{p(n-1)}\sum\limits_{q=1}^{p}\sum\limits_{i=1}^{n}(\phi_{i,q}-\overline{\phi}_q)^2 < d_{\tilde{t}}$.
We circumvent the general problem of selecting an appropriate variance threshold with an annealing approach that gradually decreases the threshold $d_{\tilde{t}}$ by a factor of $10$ every $r$ epochs.
The overall loss function used in the training of the regularized deep SVDD will be discussed in the next sub-section.

\subsection{Adaptive Weighting Scheme}\label{Adaptive weighting scheme}
We minimize the following objective function for the regularized variant of deep SVDD.
\begin{equation}
\mathcal{L}_{final}= \mathcal{L}_{SVDD} + c_t \cdot \mathcal{L}_{reg},
\label{total_loss}
\end{equation}
where $c_t$ is the adaptive weighted constant  defined as follows.
\begin{equation}
c_t = \alpha c_{t-1} +\beta(1-\alpha) \frac{\mathcal{L}_{SVDD}^{(t)}}{\mathcal{L}_{reg}^{(t)}}
\label{weighted_constant}
\end{equation}
At every iteration $t$, $c_t$ is updated based on the value of its previous iteration $c_{t-1}$ and the ratio of the SVDD loss $\mathcal{L}_{SVDD}^{(t)}$ and regularization loss $\mathcal{L}_{reg}^{(t)}$ from the current minibatch. The contribution of the different components to $c_{t}$ is determined by the hyperparameter $\alpha \in [0,1]$, while the hyperparameter $\beta \in [0,1]$ controls the ratio of the two losses. 
This adaptive weighting scheme prevents the SVDD loss from dominating the overall loss \eqref{total_loss} in the early stages of training. Moreover, it allows an appropriate amount of regularization to be applied without harshly impeding the learning of feature representations. We consider one regularization method at a time and not both methods jointly as we found them not to improve if applied together. Thus, $\mathcal{L}_{reg}$ refers to either the loss of the regularizer in Section \ref{Reg with noise inject aux} or \ref{Reg with minibatch var}. For the regularizer in Section \ref{Reg with noise inject aux}, the loss is expected to converge to $\mathbb{E}[\mathcal{L}_{reg}]= -(\frac{1}{2}\log \frac{1}{2}+ \frac{1}{2}\log \frac{1}{2})=\log 2$, as the linear layer acts essentially as a random multi-label classifier trained with binary labels sampled from the uniform discrete distribution. We observe this convergence in our experiments.

\section{Experimental Settings}\label{EXPERIMENTAL SETTINGS}
\sisetup{round-mode=places,round-precision=1,detect-all}
\subsection{Datasets}\label{dataset}
We used the CIFAR10 \cite{krizhevsky2009learning} dataset and the more challenging VOC2012 \cite{pascal-voc-2012} and WikiArt \cite{WikiArto54:online,tan2017artgan} datasets in this work. \changemarker{For reasonable computational speed, the VOC2012 and Wikiart images were resized to be on the same scale as the input samples used in pretrained CNNs which is $224 \times 224$.} For the CIFAR10 and VOC2012 datasets, all the classes were included.  For the Wikiart dataset, we included only six styles, i.e. \textit{Art Nouveau Modern}, \textit{Contemporary Realism}, \textit{Fauvism}, \textit{High Renaissance}, \textit{Naive Art Primitivism}, and \textit{Symbolism} to save training time. We refer to these original, non-manipulated images as natural images. 
We considered two different experimental setups: the common synthetic one-vs-all setup \cite{ruff2018deep,golan2018deep} and our novel setup with image alteration anomalies. In the common setup, only a single class is considered normal at a time, and the remaining $C-1$ classes are considered anomalous. 
We followed the experimental setups in \cite{ruff2018deep,ruff2019deep}. For the CIFAR10 dataset, we used 80\% of the normal class samples from the given train set as training samples, and the remaining samples in the given train set were used as the validation set for unsupervised learning.
For semi-supervised learning, we used 80\% of the normal samples from the given train set and 10\% of samples from a single anomaly class as training samples. The remaining samples were used for validation. Performance was reported on the given test set.

In our novel setup with image alteration anomalies, we pooled over all natural samples from $C$ classes and considered them as the normal sample set. We trained only a single anomaly detector using only natural samples from all the $C$ classes. To create the anomalies, we explored two different methods.
The first method involved distorting the natural image by splitting the image into $N\times N$ blocks and permuting these blocks. The second method involved drawing with certain thickness $M$, a number of random strokes (one random color per image) of different lengths on various parts of the natural image. Larger values of $N$ and $M$ indicate a larger manipulation/distortion of the natural image. The anomalies created from the first method have apparent geometrical structures while the anomalies from the second method do not. We created an anomalous counterpart for each natural sample.
For the CIFAR10 dataset, we split 20\% of the given train set for validation. We obtained 20,000 samples for both validation and test, and 40,000 training samples for unsupervised learning. 
For the VOC2012 dataset, we split 60\% of the given validation set for test. We obtained 4658 and 6988 validation and test samples respectively, 5717 training samples for unsupervised learning, and 6352 training samples for semi-supervised learning.
For the Wikiart dataset, 20\% of the given train set was split as validation and the given validation set was used as test set. We obtained 3928 and 8410 validation and test samples respectively, 7856 training samples for unsupervised learning, and 8728 training samples for semi-supervised learning.
All train sets for unsupervised learning consist of normal samples only, while the train sets for semi-supervised learning consist of $90\%$ unlabeled samples and $10\%$ labeled anomalies. There are equal number of anomalies and normal samples in both validation and test sets. 
Our novel setup is more challenging and representative of normal samples encountered in the wild as compared to the one-vs-all setup. AD on the one-vs-all setup which assumes the normal samples come only from a single class may be too easy for datasets with strong geometric shape priors. This will be discussed in Section \ref{RESULTS}. In addition, our setup adds to the variety of benchmarks and allows us to work on actual anomaly detection problems, as opposed to problems for which specific prior knowledge is available but are treated generically as anomaly detection problems \cite{xu2017feature,canali2011prophiler}. 
We used the area under the curve (AUC) of the receiver operating curve as a performance metric. Unless otherwise stated, all reported results are the average of five runs, where each run was initialized with a random seed.

\subsection{Model Hyperparameters}
We used ResNet-18 \cite{he2016deep} as the underlying network architecture with max-pooling layer. For the deep SVDD model and its variants, we finetuned the pretrained ResNet-18 network as opposed to \cite{ruff2018deep,ruff2019deep} which trained an autoencoder for network initialization. We used the Adam optimizer \cite{kingma2014adam} with $\lambda =5\cdot 10^{-4}$ and a batch size of $50$. We set 30 as the maximum number of epochs and selected the best epoch model based on the validation performance. 
For the proposed regularized variants, we set $c_0=0, \alpha=0.9$, and $\beta=0.5$. 
For the method in Section \ref{Reg with noise inject aux}, the optimal number of $k$ classification tasks to use was determined by selecting $k$ from $k \in \{5, 30, 50, 60, 70, 100, 130, 200\}$ via grid search using the validation performance as a selection criterion.  
We used $r=3$ and $d_0=0.1$ for the method in Section \ref{Reg with minibatch var}. 
A non-decaying learning rate was used to fully analyse the impact of mode collapse. We used a lower learning rate for the other layers as compared to the linear layer $g(\cdot\ ;\mathcal{W}_1)$ (if present). Unless otherwise stated, the learning rate $10^{-3}$ is used for the linear layer $g(\cdot\ ;\mathcal{W}_1)$ (if present), while the learning rate for the other layers is set to $10^{-4}$.  For
deep SVDD in \textit{soft-boundary} mode, we set $\nu=0.2$ and $0.1$ for the VOC2012 and WikiArt datasets respectively. For semi-supervised learning, we set $\eta=1.0$.
We used the PyTorch framework \cite{paszke2017automatic} for implementation.  

For the state-of-the art method \cite{golan2018deep}, which we refer to as \textit{GEO w/ Dirichlet} in Section \ref{RESULTS}, we used the Adam optimizer with $\lambda =5\cdot 10^{-4}$, a batch size of $35$, and a learning rate of $10^{-4}$. As suggested, we trained the model for $\lceil200/|\mathcal{T}|\rceil$ epochs, where $\mathcal{T}$ is the set of $72$ transformations. To control architectural effects, we used the pretrained ResNet-18 as the underlying architecture instead of the Wide Residual Network \cite{zagoruyko2016wide} as done in \cite{golan2018deep}. The model was trained with the Keras framework \cite{chollet2015keras} with TensorFlow as the backend implementation. For the shallow models such as OC-SVM \cite{scholkopf2000support} and IF \cite{liu2008isolation}, we used as inputs either features from PCA of the raw data, or features extracted from pretrained ResNet-18 after the max-pooling layer. We selected the same set of hyperparameters as mentioned in \cite{liu2008isolation,erfani2016high,ruff2018deep}.

\section{Results and Discussion}\label{RESULTS}
\sisetup{round-mode=places,round-precision=1,detect-all}
\begin{table}
\begin{center}
\renewcommand{\arraystretch}{1.2}
\setlength\tabcolsep{2.5pt} 
{\caption{Effect of adaptive weighting scheme on the proposed regularizers. Methods that do not adopt the adaptive weighting scheme adopt the fixed weighting scheme of $c_t=1, \forall t$. Performance is reported in \textit{one-class} mode. }\label{adaptive_weight_vs_non}}
\begin{tabular}{lcllll}
\hline
 & \multicolumn{1}{c}{} & \multicolumn{2}{c}{\textbf{VOC2012  Dataset}} & \multicolumn{2}{c}{\textbf{WikiArt Dataset}} \\ \cline{3-6} 
\multirow{-2}{*}{\textbf{Models}} & \multicolumn{1}{c}{\multirow{-2}{*}{\textbf{\begin{tabular}[c]{@{}c@{}}Adaptive\\Weighting\end{tabular}}}} & \multicolumn{1}{c}{\textit{Strokes $5$}} & \multicolumn{1}{c}{\textit{Strokes $9$}} & \multicolumn{1}{c}{\textit{Strokes $5$}} & \multicolumn{1}{c}{\textit{Strokes $9$}} \\ \hline\hline
 & \xmark &\num{75.0}$\pm$\num{1.1}  &\num{76.2}$\pm$\num{4.4}  & \num{81.4}$\pm$\num{1.3}  & \num{84.0}$\pm$\num{1.3}  \\
\multirow{-2}{*}{\begin{tabular}[c]{@{}c@{}}D. SVDD w/ bias \&\\noise reg. \end{tabular}} & \cmark \cellcolor[HTML]{EFEFEF} & \cellcolor[HTML]{EFEFEF}\textbf{\num{76.6}$\pm$\num{1.8}} & \cellcolor[HTML]{EFEFEF}\textbf{\num{78.8}$\pm$\num{1.0}} & \cellcolor[HTML]{EFEFEF}\textbf{\num{81.8}$\pm$\num{1.2}} & \cellcolor[HTML]{EFEFEF}\textbf{\num{84.2}$\pm$\num{1.3}}\\  \hline
 & \xmark  &\num{73.1}$\pm$\num{2.6} &\num{77.3}$\pm$\num{1.9}  & \textbf{\num{78.1}$\pm$\num{6.1}} &\textbf{\num{84.6}$\pm$\num{0.4}}  \\

\multirow{-2}{*}{\begin{tabular}[c]{@{}c@{}}D. SVDD w/ bias \&\\$\sigma^2$ reg. \end{tabular}} & \cmark \cellcolor[HTML]{EFEFEF} & \cellcolor[HTML]{EFEFEF}\textbf{\num{78.7}$\pm$\num{3.2}} & \cellcolor[HTML]{EFEFEF}\textbf{\num{80.8}$\pm$\num{2.3}} & \cellcolor[HTML]{EFEFEF}\num{75.7}$\pm$\num{3.6} & \cellcolor[HTML]{EFEFEF}\num{78.5}$\pm$\num{1.5}\\ \hline
\end{tabular}
\end{center}
\end{table}
\begin{table}
\begin{center}
\renewcommand{\arraystretch}{1.2}
\setlength\tabcolsep{3.5pt} 
{\caption{Performance on the \textit{block permutation} anomaly test sets. Performance of the deep SVDD and its variants is reported in \textit{one-class} mode. }\label{block_perm_strokes}} 
\begin{tabular}{clllll} 
\hline
\multicolumn{2}{c}{\multirow{2}{*}{\textbf{Models} }}                                                                                                               & \multicolumn{2}{c}{\textbf{VOC2012 Dataset} }                                                                                                                                                                                           & \multicolumn{2}{c}{\textbf{WikiArt Dataset} }                                                                                                                                                                                            \\ 
\cline{3-6}
\multicolumn{2}{c}{}                                                                                                                                                & \multicolumn{1}{c}{\begin{tabular}[c]{@{}c@{}}\textit{Block}\\\textit{ perm.}\\\textit{ $2\times2$} \end{tabular}} & \multicolumn{1}{c}{\begin{tabular}[c]{@{}c@{}}\textit{Block }\\\textit{perm. }\\\textit{$8\times8$} \end{tabular}} & \multicolumn{1}{c}{\begin{tabular}[c]{@{}c@{}}\textit{Block}\\\textit{ perm. }\\\textit{$2\times2$} \end{tabular}} & \multicolumn{1}{c}{\begin{tabular}[c]{@{}c@{}}\textit{Block }\\\textit{perm. }\\\textit{$8\times8$} \end{tabular}}  \\ 
\hline\hline
\multicolumn{2}{c}{OC-SVM w/ PCA}&
 \num{59.7524302844454}$\pm$\num{0.000246334290557158}                                                                         & \num{59.791817455322}$\pm$\num{0.000363176079601644} & \num{55.9551782106404}$\pm$\num{0.000184218769907023 }                                                                & \num{53.1604134707421}$\pm$\num{0.00015327655298475}                                                                            \\                                                                          \multicolumn{2}{c}{\cellcolor[rgb]{0.937,0.937,0.937}OC-SVM w/ CNN feat.}                               & {\cellcolor[rgb]{0.937,0.937,0.937}}\num{53.457650354242}$\pm$\num{0.0000143289679696143}                                      & {\cellcolor[rgb]{0.937,0.937,0.937}}\num{52.906595295099}$\pm$\num{0.0000125837534342992}                                      & {\cellcolor[rgb]{0.937,0.937,0.937}}\num{51.7200665647741}$\pm$\num{0.000241981037826521}                                      & {\cellcolor[rgb]{0.937,0.937,0.937}}\num{64.3714088177117}$\pm$\num{0.0515182570126213}                                         \\
                                                                                  \multicolumn{2}{c}{IF w/ PCA}                                                                            & \num{59.5520344140531}$\pm$\num{0.287694356623363}                                                                             & \num{82.7982011193282}$\pm$\num{0.218769178601452}                                                                             & \num{59.2035595470541}$\pm$\num{0.103021620343201}                                                                             & \num{74.7466152773791}$\pm$\num{0.260995827210997}                                                                              \\
                                                                          \multicolumn{2}{c}{\cellcolor[rgb]{0.937,0.937,0.937}IF w/ CNN feat.}                                    & {\cellcolor[rgb]{0.937,0.937,0.937}}\num{51.6565875133395}$\pm$\num{0.726356534628857}                                         & {\cellcolor[rgb]{0.937,0.937,0.937}}\num{30.8492324236265}$\pm$\num{3.82195042589981}                                         & {\cellcolor[rgb]{0.937,0.937,0.937}}\num{50.8593670690998}$\pm$\num{1.09550339747512}                                          & {\cellcolor[rgb]{0.937,0.937,0.937}}\num{38.6611239380104}$\pm$\num{2.01622225031989}                                           \\
                                                                          \multicolumn{2}{c}{GEO w/ Dirichlet \cite{golan2018deep}} &\textbf{\num{78.7940648274628}$\pm$\num{0.527561661327793}} & \textbf{\num{87.400461466529}$\pm$\num{1.11597319201243}} &\textbf{\num{81.1115310604978}$\pm$\num{1.09684975368268}} & \textbf{\num{88.3100810003379}$\pm$\num{1.93191453821904}} \\

\hline
\multicolumn{2}{c} {D. SVDD\cite{ruff2018deep}}                                                                       & \num{65.5925760703851}$\pm$\num{1.65512346691851}                                                                              & \num{80.5776981653723}$\pm$\num{2.75139752943381}                                                                              & \num{69.0293696564731}$\pm$\num{0.497503744709877}                                                                             & \num{78.9448493597311}$\pm$\num{ 2.28134175210143}                                                                               \\
                                                                        {\cellcolor[rgb]{0.937,0.937,0.937}}  & {\cellcolor[rgb]{0.937,0.937,0.937}}D. SVDD w/ bias                                  & {\cellcolor[rgb]{0.937,0.937,0.937}}\num{65.9691837409391}$\pm$\num{1.05700536517539}                                         & {\cellcolor[rgb]{0.937,0.937,0.937}}\num{79.0345654288699}$\pm$\num{3.57896041159125}                                          & {\cellcolor[rgb]{0.937,0.937,0.937}}\num{69.0510328992295}$\pm$\num{0.66243546154259}                                          & {\cellcolor[rgb]{0.937,0.937,0.937}}\num{79.0324388750722}$\pm$\num{1.04797249363812}                                           \\
                                                                          & \begin{tabular}[c]{@{}l@{}}D. SVDD w/ bias \&\\noise reg. \textit{(proposed)} \end{tabular} & \num{66.7238497658427}$\pm$\num{1.32116091594531}                                                                            & \num{81.7356338071086}$\pm$\num{3.49829938735858}                                                                              & \num{69.6570064797442} $\pm$\num{0.488541480599869}                                                                             & \num{79.7361829315364}$\pm$\num{0.857786634117532}                                                                              \\
                                                 {\cellcolor[rgb]{0.937,0.937,0.937}}                            & {\cellcolor[rgb]{0.937,0.937,0.937}}\begin{tabular}[c]{@{}l@{}}D. SVDD w/ bias \&\\$\sigma^2$ reg. \textit{(proposed)} \end{tabular} & {\cellcolor[rgb]{0.937,0.937,0.937}}\num{66.27706454994071}$\pm$\num{1.4481435254885945}                                                                              & {\cellcolor[rgb]{0.937,0.937,0.937}}\num{79.44057995897129}$\pm$\num{2.1725658696836094}                                                                              & {\cellcolor[rgb]{0.937,0.937,0.937}}\num{67.69078767844745}$\pm$\num{1.0718953562655228}                                                                            & {\cellcolor[rgb]{0.937,0.937,0.937}}\num{80.699036450859}$\pm$\num{3.9304339866227896}
                                                \\                             
\hline
\end{tabular}
\end{center}
\end{table}

In this section, we study the effect of the adaptive weighting scheme on the regularizers, followed by a comparison of the different methods on the two setups (refer Section \ref{dataset}). In addition, we explain the stability and performance at different learning rates and discuss the impact of the pooling layer on performance. Lastly, we also extend our experiments to include the semi-supervised deep SVDD. Note that in all tables, we refer to the noise-regularized variant described in Section \ref{Reg with noise inject aux} as \textit{D. SVDD w/ bias \& noise reg.} and the variance-regularized method described in Section \ref{Reg with minibatch var} as \textit{D. SVDD w/ bias \& $\sigma^2$ reg.}. To study the effect of the proposed regularizers on performance, we also include the comparison with a non-regularized variant of deep SVDD with the bias term, i.e. \textit{D. SVDD w/ bias}. This variant which uses the bias term is prone to mode collapse and is not to be confused with the standard deep SVDD model \cite{ruff2018deep} that follows a set of architectural constraints. We refer to the standard deep SVDD model as \textit{D. SVDD} in the tables. In Table \ref{adaptive_weight_vs_non}, it can be observed that the adaptive weighting scheme generally improves performance in a majority of the settings. Therefore, for the remaining experiments, we use only the adaptive weighting scheme and refrain from using the fixed weighting scheme.
\begin{table}
\begin{center}
\renewcommand{\arraystretch}{1.2}
\setlength\tabcolsep{10pt} 
{\caption{Performance on \textit{one-vs-all} and \textit{strokes} anomaly test sets for the CIFAR10 dataset. Results for \textit{one-vs-all} are the average of $10$ normal classes. Performance of the deep SVDD and its variants is reported in \textit{one-class} mode.}\label{one_vs_all_strokes1}}
\begin{tabular}{cll}
\hline
 & \multicolumn{2}{c}{\textbf{CIFAR10 Dataset}} \\\cline{2-3} 
\multirow{-2}{*}{\textbf{Models}} & \multicolumn{1}{c}{\textit{One-vs-all}} & \multicolumn{1}{c}{\textit{Strokes $1$}} \\
\hline\hline
OC-SVM w/ PCA & \num{61.6}$\pm$\num{10.37732592} & \num{53.6}$\pm$\num{0.0} \\
\rowcolor[HTML]{EFEFEF} 
OC-SVM w/ CNN feat. & \num{75.3}$\pm$\num{8.045347715} & \num{66.7}$\pm$\num{0.0} \\
IF w/ PCA & \num{54.5}$\pm$\num{12.56585076} & \num{64.1}$\pm$\num{0.8} \\
\rowcolor[HTML]{EFEFEF} 
IF w/ CNN feat. & \num{69.2}$\pm$\num{10.04525908}  & \num{50.9}$\pm$\num{1.7} \\
GEO w/ Dirichlet \cite{golan2018deep} &\textbf{\num{87.2}$\pm$\num{7.204095652} } & \num{73.7}$\pm$\num{1.5} \\
\hline
\multicolumn{1}{l}{D. SVDD\cite{ruff2018deep}} & \num{67.85}$\pm$\num{6.513936724} & \num{85.7}$\pm$\num{7.1} \\
\rowcolor[HTML]{EFEFEF} 
\multicolumn{1}{l}{\cellcolor[HTML]{EFEFEF}D. SVDD w/ bias} &\num{67.7}$\pm$\num{6.125899241}  & \num{85.1}$\pm$\num{7.5}  \\
\multicolumn{1}{l}{\begin{tabular}[c]{@{}l@{}}D. SVDD w/ bias \&\\noise reg. \textit{(proposed)}\end{tabular}} &\num{68.6}$\pm$\num{6.432439757}  & \textbf{\num{91.3}$\pm$\num{3.6} }\\
\rowcolor[HTML]{EFEFEF} 
\multicolumn{1}{l}{\cellcolor[HTML]{EFEFEF}\begin{tabular}[c]{@{}l@{}}D. SVDD w/ bias \&\\$\sigma_2$ reg. \textit{(proposed)}\end{tabular}} &\num{67.6}$\pm$\num{5.191544255}  & \num{86.0}$\pm$\num{5.0}\\
\hline
\end{tabular}
\end{center}
\end{table}

\begin{figure}[!ht]
\centerline{\includegraphics[height=1.2in, width=2.9in]{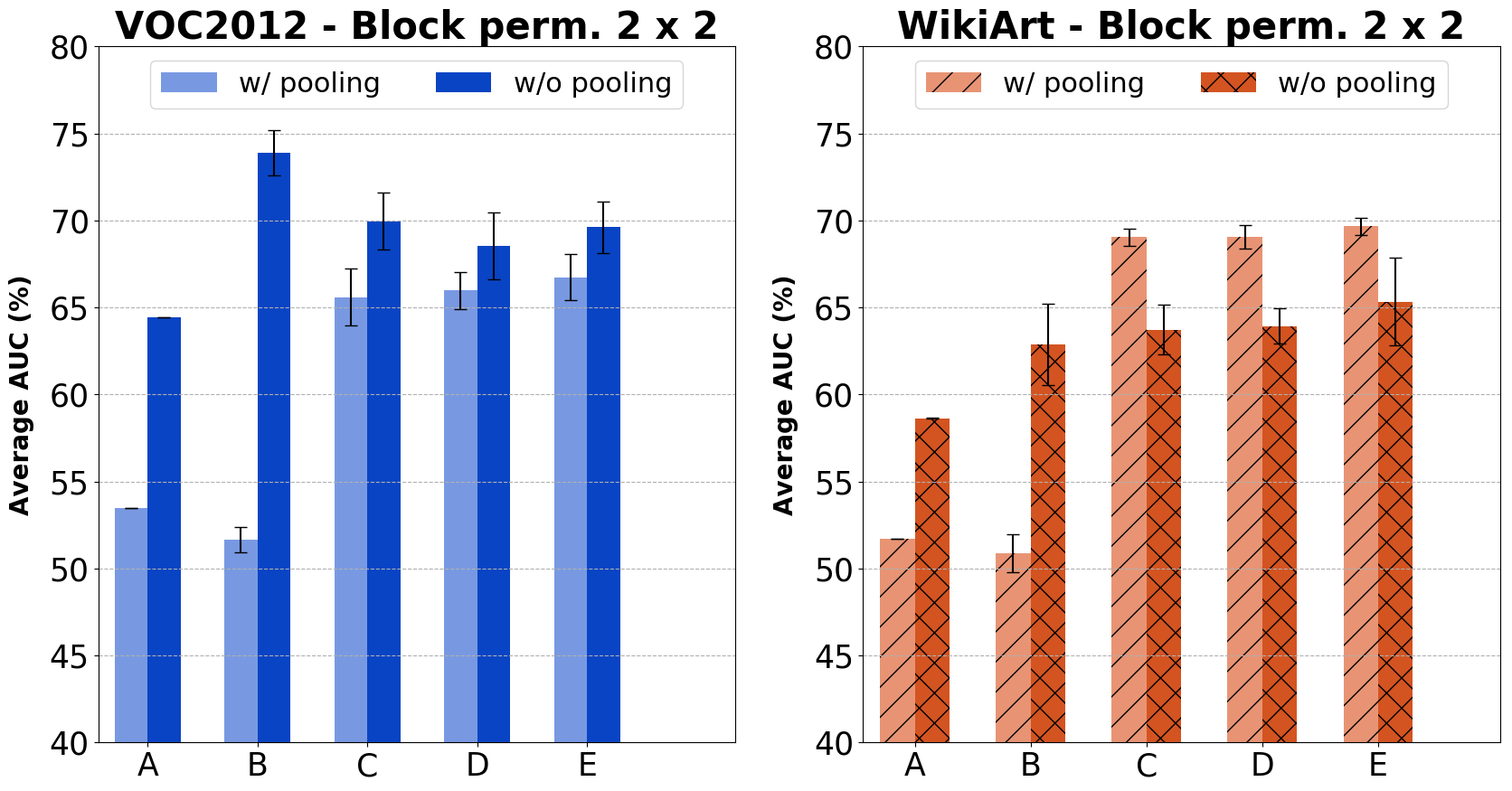}}
\caption{Performance of models with and without the pooling layer in ResNet-18. The deep SVDD model and its variants are trained with the \textit{one-class} objective function. We denote the OC-SVM, IF, standard deep SVDD \cite{ruff2018deep}, deep SVDD with bias term, and noise-regularized deep SVDD models as A, B, C, D, and E, respectively.} \label{w_wo_pooling}
\end{figure}

\begin{table}
\begin{center}
\renewcommand{\arraystretch}{1.2}
\setlength\tabcolsep{2.5pt} 
{\caption{Performance on the \textit{strokes} anomaly test sets.}\label{strokes}} 
\begin{tabular}{clllll}
\hline
\multicolumn{2}{c}{\multirow{2}{*}{\textbf{Models}}} &  \multicolumn{2}{c}{\textbf{VOC2012 Dataset}} & \multicolumn{2}{c}{\textbf{WikiArt Dataset}} \\ \cline{3-6} 
\multicolumn{2}{c}{} & \multicolumn{1}{c}{\textit{Strokes $5$}} & \multicolumn{1}{c}{\textit{Strokes $9$}} & \multicolumn{1}{c}{\textit{Strokes $5$}} & \multicolumn{1}{c}{\textit{Strokes $9$}} \\ \hline
\hline
\multicolumn{2}{c}{OC-SVM w/ PCA} &\num{50.7833069954905}$\pm$\num{0.000162394970351841} &\num{56.1382461519608}$\pm$\num{0.0000968033122613926} &\num{55.4869411167555}$\pm$\num{0.000442561402317869}&\num{62.5275040613278}$\pm$\num{0.000338522677723001}  \\  
 \multicolumn{2}{c}{\cellcolor[HTML]{EFEFEF}OC-SVM w/ CNN feat.} &\cellcolor[HTML]{EFEFEF}\num{54.3806219116653}$\pm$\num{0.0000867043224471323}  &\cellcolor[HTML]{EFEFEF}\num{54.6100232666417}$\pm$\num{0.0000131061212499394}  &\cellcolor[HTML]{EFEFEF}\num{57.3915793015788}$\pm$\num{0.000208987452865645}  &\cellcolor[HTML]{EFEFEF}\num{55.2140249773428}$\pm$\num{0.000250943191834473}  \\ 
 \multicolumn{2}{c}{IF w/ PCA} &\num{68.4498964452595}$\pm$\num{0.537736861742241}  &\num{74.3124020931786}$\pm$\num{0.454394919842671}  &\num{68.5468332953946}$\pm$ \num{0.18448211505888} & \num{74.0593512338094}$\pm$\num{0.301200972540863} \\  
 \multicolumn{2}{c}{\cellcolor[HTML]{EFEFEF}IF w/ CNN feat.} &\cellcolor[HTML]{EFEFEF}\num{41.914815781998}$\pm$\num{2.02946929538577} &\cellcolor[HTML]{EFEFEF}\num{38.949718038184}$\pm$\num{3.44089019818382} &\cellcolor[HTML]{EFEFEF}\num{47.4516148461502}$\pm$\num{3.05986701228273}  &\cellcolor[HTML]{EFEFEF}\num{46.702675739911}$\pm$\num{2.7365001175189}  \\
 \multicolumn{2}{c}{GEO w/ Dirichlet \cite{golan2018deep}} &\num{66.7485728253094}$\pm$\num{2.1163460069019} & \num{70.8166702653891}$\pm$\num{0.988853742555014}&\num{63.9529850229258}$\pm$\num{0.797328957160356} &\num{69.2417152447188}$\pm$\num{1.00764405301185} \\
 \hline
\multirow{4}{*}{\begin{tabular}[c]{@{}l@{}}Soft-\\ bound.\end{tabular}} & D. SVDD\cite{ruff2018deep} &\num{74.1632847412966}$\pm$\num{3.13468145549769}  &\num{75.4094065581064}$\pm$\num{2.48750782772519}  & \textbf{\num{80.3522492474702}$\pm$\num{0.529989386359478}} & \num{83.5492716473367}$\pm$\num{0.61265848355125}\\ 
 &\cellcolor[HTML]{EFEFEF}D. SVDD w/ bias & \cellcolor[HTML]{EFEFEF}\num{71.5985421406031}$\pm$\num{2.84353960590467} &\cellcolor[HTML]{EFEFEF}\num{74.4609485096538}$\pm$\num{4.14144665215027}& \cellcolor[HTML]{EFEFEF}\num{79.9663070264859}$\pm$\num{0.983326777378554} & \cellcolor[HTML]{EFEFEF}\num{82.9990382888837}$\pm$\num{0.810596635625827} \\  
     &\begin{tabular}[c]{@{}l@{}}D. SVDD w/ bias \&\\noise reg. \textit{(proposed)}\end{tabular} &\textbf{\num{75.544858321191}$\pm$\num{1.90218143592121}}  & \num{76.7077325132396}$\pm$\num{2.00872877131784}  &\textbf{\num{80.395589306089}$\pm$\num{0.977540767724493}}  & \textbf{\num{84.2301863050188}$\pm$\num{1.84188989082894}} \\ 
       & {\cellcolor[rgb]{0.937,0.937,0.937}}\begin{tabular}[c]{@{}l@{}}D. SVDD w/ bias \&\\$\sigma^2$ reg. \textit{(proposed)} \end{tabular} & {\cellcolor[rgb]{0.937,0.937,0.937}}\num{74.78299949312077}$\pm$\num{6.326796152642004}                                                                              & {\cellcolor[rgb]{0.937,0.937,0.937}}\textbf{\num{79.78422081979444}$\pm$\num{1.637013038460576}}                                                                              & {\cellcolor[rgb]{0.937,0.937,0.937}}\num{75.26494222239817}$\pm$\num{3.0278048520597223}                                                                            & {\cellcolor[rgb]{0.937,0.937,0.937}}\num{77.55393231261691}$\pm$\num{3.0131935645808516} \\
     
     \hline
\multirow{4}{*}{\begin{tabular}[c]{@{}l@{}}One-\\ class\end{tabular}} & D. SVDD\cite{ruff2018deep} &\num{73.559538979079}$\pm$\num{4.71487768411646}  &\num{70.2172003752282}$\pm$\num{4.48611921817584}  &\textbf{\num{81.8425830751851}$\pm$\num{0.70824999682687}}  &\num{84.0422412025772}$\pm$\num{0.778809122956158}  \\  
 &\cellcolor[HTML]{EFEFEF}D. SVDD w/ bias &\cellcolor[HTML]{EFEFEF}\num{72.9048005756208}$\pm$\num{5.04281267807125}  &\cellcolor[HTML]{EFEFEF}\num{76.511438039665}$\pm$\num{2.66877193307964} &\cellcolor[HTML]{EFEFEF}\num{81.3662066420559}$\pm$\num{0.850576700515012}  &\cellcolor[HTML]{EFEFEF}\num{83.4277193929993}$\pm$\num{2.85354546556411} \\ 
 & \begin{tabular}[c]{@{}l@{}}D. SVDD w/ bias \&\\noise reg. \textit{(proposed)}\end{tabular} &\num{76.6492292453921}$\pm$\num{1.75511915786852} &\num{78.7890836822565}$\pm$\num{1.00291812138226} &\textbf{\num{81.8452224787602}$\pm$\num{1.1627550537664}}  & \textbf{\num{84.1699087067233}$\pm$\num{1.27485292593957}} \\ 
     & {\cellcolor[rgb]{0.937,0.937,0.937}}\begin{tabular}[c]{@{}l@{}}D. SVDD w/ bias \&\\$\sigma^2$ reg. \textit{(proposed)} \end{tabular} & {\cellcolor[rgb]{0.937,0.937,0.937}}\textbf{\num{78.74764130774189}$\pm$\num{3.222894599431399} }                                                                             & {\cellcolor[rgb]{0.937,0.937,0.937}}\textbf{\num{80.827039664693}$\pm$\num{2.2896503472585133}}                                                                              & {\cellcolor[rgb]{0.937,0.937,0.937}}\num{75.71769240231252}$\pm$\num{3.5556892124933626}                                                                            & {\cellcolor[rgb]{0.937,0.937,0.937}}\num{78.52303115734764}$\pm$\num{1.4612011767549834} \\
 \hline
\end{tabular}
\end{center}
\end{table}
\begin{table}
\begin{center}
\renewcommand{\arraystretch}{1.2}
\setlength\tabcolsep{2pt} 
{\caption{Performance of the deep SVDD model and its variants across different learning rates in \textit{soft-boundary} mode. The learning rate specified refers to the learning rate of the linear layer $g(\cdot\ ;\mathcal{W}_1)$ (if present), while the other layers in the network used a learning rate that is lower by a factor of 10 than the specified learning rate.}\label{sb_unsupervised_df_lr}}
\begin{tabular}{lllllll}
\hline
\multicolumn{1}{c}{} & \multicolumn{6}{c}{\textbf{VOC2012 Dataset}} \\ \cline{2-7}
\multicolumn{1}{c}{} & \multicolumn{3}{c}{\textit{\textbf{Block perm. 2 $\times$ 2}}} & \multicolumn{3}{c}{\textit{\textbf{Block perm. 8 $\times$ 8}}} \\ \cline{2-7} 
\multicolumn{1}{c}{\multirow{-3}{*}{\textbf{Models}}} & \textit{ $10^{-3}$} & \textit{ $10^{-2}$} & \textit{ $10^{-1}$} & \textit{ $10^{-3}$} & \textit{ $10^{-2}$} & \textit{ $10^{-1}$} \\ \hline\hline
D. SVDD\cite{ruff2018deep} &\num{65.4212503960506}$\pm$\num{1}& \textbf{\num{65.9278904485537}$\pm$\num{4}}& \num{56.5513822206946}$\pm$\num{3} & \num{76.0704629311381}$\pm$\num{2} & \num{67.3}$\pm$\num{7} & \num{57.9734160351427}$\pm$\num{5}  \\
\rowcolor[HTML]{EFEFEF} 
D. SVDD w/ bias &\num{66.5022744035158}$\pm$\num{1}& \num{64.245414250089}$\pm$\num{1}&\num{57.0560825672532}$\pm$\num{2}  &\num{76.2001389904158}$\pm$\num{2} & \num{66.6282012929844}$\pm$\num{7} &\num{64.2709113898419}$\pm$\num{4} \\
\begin{tabular}[c]{@{}l@{}}D. SVDD w/ bias \&\\noise reg. \textit{(proposed)}\end{tabular} & \num{66.0242761407322}$\pm$\num{2}& \num{65.5783403653134}$\pm$\num{2}&\textbf{\num{59.7001606155158}$\pm$\num{1}}  &\textbf{\num{78.3780503268502}$\pm$\num{4}}  & \textbf{\num{73.0680815489076}$\pm$\num{3}}&\textbf{\num{65.3703363915375}$\pm$\num{4}} \\
\rowcolor[HTML]{EFEFEF} 
\begin{tabular}[c]{@{}l@{}}D. SVDD w/ bias \&\\$\sigma^2$ reg. \textit{(proposed)}\end{tabular} & \textbf{\num{66.7152972025967}$\pm$\num{3}}& \num{64.0252477957961}$\pm$\num{2}&\num{58.4733473918327}$\pm$\num{3}  &\num{75.4239371509061}$\pm$\num{2} & \num{59.8553387293418}$\pm$\num{8}&\num{60.9309572809254}$\pm$\num{9} \\
\hline
\multicolumn{1}{c}{} & \multicolumn{6}{c}{\textbf{WikiArt Dataset}} \\ \hline\hline

D. SVDD\cite{ruff2018deep} &\num{70.1432296357459}$\pm$\num{1}& \num{59.882264050639}$\pm$\num{4}& \num{54.5566313813038}$\pm$\num{2} & \num{78.1635186015176}$\pm$\num{1} & \num{70.31180082598}$\pm$\num{5} & \num{57.8332532614335}$\pm$\num{5}  \\
\rowcolor[HTML]{EFEFEF} 
D. SVDD w/ bias &\num{69.6856044485855}$\pm$\num{1}& \num{62.1708707006126}$\pm$\num{6}&\num{58.6876259930635}$\pm$\num{1}  & \num{80.2149561489704}$\pm$\num{2}& \num{63.3590021504889}$\pm$\num{6} & \num{56.4121569786266}$\pm$\num{5} \\
\begin{tabular}[c]{@{}l@{}}D. SVDD w/ bias \&\\noise reg. \textit{(proposed)}\end{tabular} & \textbf{\num{70.2364893161275}$\pm$\num{1}}& \textbf{\num{64.8814024411796}$\pm$\num{4}}&\textbf{\num{59.5202268970889}$\pm$\num{3}}  & \textbf{\num{81.5294933696791}$\pm$\num{3}} & \textbf{\num{72.9633528965149}$\pm$\num{6}}&  \textbf{\num{62.2372352714126}$\pm$\num{6}}\\
\rowcolor[HTML]{EFEFEF} 
\begin{tabular}[c]{@{}l@{}}D. SVDD w/ bias \&\\$\sigma^2$ reg. \textit{(proposed)}\end{tabular} & \num{67.7906551992772}$\pm$\num{2}& \num{60.61994822425599}$\pm$\num{4}&\num{55.6480188213736}$\pm$\num{2}  &\num{76.1083903003191}$\pm$\num{3} & \num{64.72827235568325}$\pm$\num{10}&\num{57.2912717858955}$\pm$\num{1} \\\hline
\end{tabular}
\end{center}
\end{table}
\begin{figure}
\centerline{\includegraphics[width=3.0in, height=3.2in]{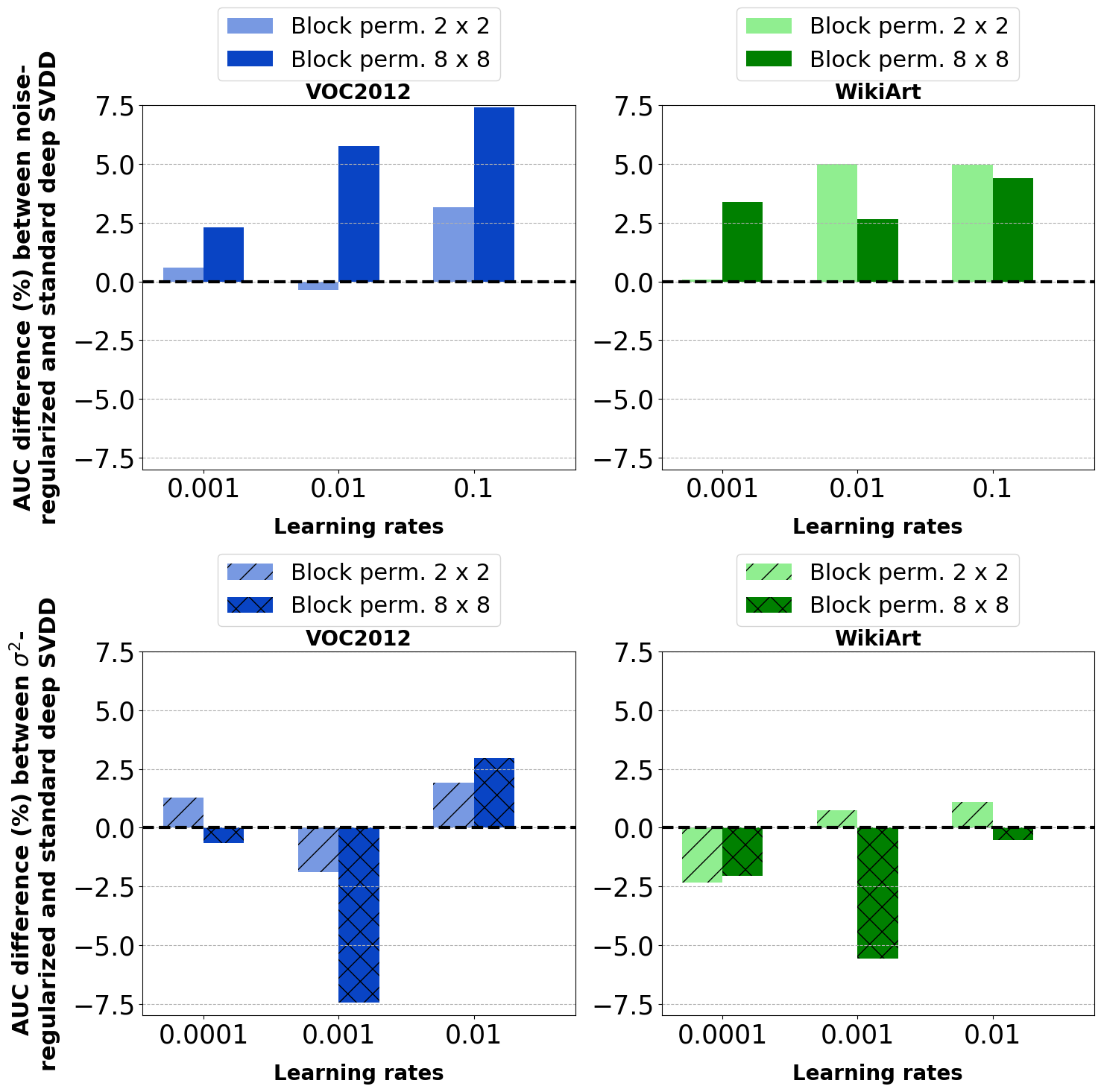}}
\caption{A positive performance difference indicates that the proposed variant outperforms the standard deep SVDD model. Models are trained with the \textit{soft-boundary} objective function. For the top row, the learning rates are those of the linear layer $g(\cdot\ ;\mathcal{W}_1)$, for all other layers it is lower by a factor of 10. 
The learning rates in the bottom row are used in all layers of the network.} \label{perf_gained}
\end{figure}
Table \ref{block_perm_strokes} compares the performance of models on the \textit{block permutation} test sets which contain normal samples and anomalies generated with method 1 (refer Section \ref{dataset}). On the \textit{block permutation} test sets, the GEO model \cite{golan2018deep} significantly outperforms all the other methods. The same can also be observed for the one-vs-all setup in Table \ref{one_vs_all_strokes1}. The anomalies in the \textit{block permutation} test set and the normal samples (from a single class) in the one-vs-all setup contain specific geometrical structures that are highly distinguishable from their counterparts. Therefore, the GEO model trained to distinguish salient geometrical features of transformed images performs well on these samples due to the presence of significant geometrical structures.
However, the GEO model does not perform well given anomalies that do not portray any apparent geometrical structure such as the \textit{strokes} anomalies created using method 2 in Section \ref{dataset}. We elaborate on this later with reference to Table \ref{one_vs_all_strokes1} and \ref{strokes}. In Table \ref{block_perm_strokes}, we also note that at least one of the proposed regularized variants of deep SVDD (mostly the noise-regularized variant, i.e. \textit{D. SVDD w/ bias \& noise reg.}) has the second best performance after the GEO method for a majority of the \textit{block permutation} settings. In Table \ref{one_vs_all_strokes1}, the OC-SVM model with CNN features performs second best on the one-vs-all setup. Additionally, we observe that the noise-regularized model appears to be more effective than the variance-regularized model in a majority of the cases. 
For some experiments, we also observe that the shallow models with features from PCA yield better performance than features from ResNet-18, which is partially due to the pooling operation in the pretrained model that has removed some useful features of the anomalies. This will be shown in Figure \ref{w_wo_pooling} and discussed later. \changemarker{We also note that the performance achieved by the different models for the one-vs-all setup on CIFAR10 in Table \ref{one_vs_all_strokes1} is similar to other works \cite{ruff2018deep,golan2018deep}.}

The GEO method does not perform well on the \textit{strokes} test sets as shown in Table \ref{one_vs_all_strokes1} and \ref{strokes}. Since there is no apparent geometrical structure in the \textit{strokes} test sets due to the various shapes and lengths of the strokes, the GEO model can not take advantage of the geometrical structure of the anomalies and does not perform well in this setting. This suggests that the use of the GEO model may not be appropriate for other types of anomalies that do not portray any apparent geometrical structure. For the shallow models and all variants of deep SVDD, their detections do not directly rely on the presence of geometrical structure in the anomalies and hence, can be used generically for anomaly detection regardless of the types of anomalies. Based on Table \ref{one_vs_all_strokes1} and \ref{strokes}, the variance-regularized model performs better than the other models on the VOC2012 dataset for a majority of the settings, while the noise-regularized model is more effective on the CIFAR10 and WikiArt dataset. It can also be observed that the deep SVDD model and its variants perform better on easier settings, i.e. samples with larger distortion of the natural image. 

Regarding impact of learning rates, we observe that both regularized and non-regularized deep SVDD models generally perform better at low learning rate ($0.001$) as compared to high learning rates ($0.01$ and $0.1$). 
Close examinations on the experiments also reveal that both regularized variants and the standard deep SVDD model \cite{ruff2018deep} collapse to a single point $\phi \equiv c $ at high learning rates ($0.01$ and $0.1$). For the variance-regularized method, we observe in our preliminary experiments that fixing the variance threshold $d_{\tilde{t}}$ to a sufficiently small constant can effectively prevent mode collapse for high learning rates, but such a stringent threshold can severely hinder feature learning. Thus, we choose to adopt the annealing approach to gradually decrease the variance threshold $d_{\tilde{t}}$ instead.
All results reported at high learning rates ($0.01$ and $0.1$) in Figure 
\ref{perf_gained} and Table \ref{sb_unsupervised_df_lr}, are the test performances of the best performing model before the hypersphere collapses. 
Although both regularized models and the standard deep SVDD collapse under high learning rates, \textit{D. SVDD w/ bias \& noise reg.} model achieves better performance convergence than the  \textit{D. SVDD} model for a majority of the experiments as shown in Table \ref{sb_unsupervised_df_lr}. Moreover, even though the \textit{D. SVDD w/ bias \& $\sigma^2$ reg.} model is less effective than the \textit{D. SVDD w/ bias \& noise reg.} model, it shows better performance convergence than the \textit{D. SVDD} model for half of the experiments at high learning rates ($0.01$ and $0.1$). These observations suggest that the regularizers may have curbed the impact of mode collapse to an extent. 
Figure \ref{perf_gained} clearly depicts the trend of the performance difference between the proposed methods and the standard deep SVDD across different learning rates. For the noise-regularized variant of deep SVDD, we observe in general a higher performance gain over the standard deep SVDD model as the learning rate increases. 
Meanwhile, no obvious trend is shown for the performance difference between the variance-regularized model and the standard deep SVDD model. 
For all other experiments, we choose the low learning rate of $10^{-3}$ due to better performance and stability against mode collapse.

In addition, we remove the pooling layer of ResNet-18 and compare the performance of the models with the results from Table \ref{block_perm_strokes}. In Figure \ref{w_wo_pooling}, the OC-SVM and IF models generally show better performance when given ResNet-18 features without pooling. For some settings, these shallow models even outperform their counterparts with PCA features (refer Table \ref{block_perm_strokes}). The improvement in performance suggests that the pooling operation may have removed certain useful features for the anomalous samples. Without the pooling layer, the deep SVDD model and its variants consistently perform better on the VOC2012 dataset and worse for the WikiArt dataset.
\begin{table}[h!]
\begin{center}
\renewcommand{\arraystretch}{1.2}
\setlength\tabcolsep{2.2pt} 
{\caption{Performance of the semi-supervised deep SVDD model and its variants in \textit{one-class} mode. Performance for \textit{one-vs-all} setup is the average of $10$ normal classes. For each normal class, only a single anomaly class (out of nine) was included during training and each anomaly class setting was ran five times, resulting in a total of $9\times 5$ runs.}\label{Semi-supervised_one-class_tab}}
\begin{tabular}{lcllllll}
\hline
\multicolumn{1}{c}{} &
\multicolumn{1}{c}{\textbf{CIFAR10}}&
\multicolumn{3}{c}{\textbf{VOC2012}} & \multicolumn{3}{c}{\textbf{WikiArt}} \\ \cline{2-8} 
\multicolumn{1}{c}{\multirow{-2}{*}{\textbf{Models}}}& \textit{\begin{tabular}[c]{@{}l@{}}One\\-vs-\\all\end{tabular}}& \textit{\begin{tabular}[c]{@{}l@{}}$\Diamond$Bp.$8$\\$\Box$St.$9$\\ $\bigtriangleup$St.$5$\end{tabular}} & \textit{\begin{tabular}[c]{@{}l@{}}$\Diamond$St.$9$\\ $\Box$Bp.$8$\\ $\bigtriangleup$Bp.$2$\end{tabular}} & \textit{\begin{tabular}[c]{@{}l@{}}$\Diamond$St.$9$\\ $\Box$St.$5$\\ $\bigtriangleup$Bp.$2$\end{tabular}} & \textit{\begin{tabular}[c]{@{}l@{}}$\Diamond$Bp.$8$\\ $\Box$St.$9$\\ $\bigtriangleup$St.$5$\end{tabular}} & \textit{\begin{tabular}[c]{@{}l@{}}$\Diamond$St.$9$\\ $\Box$Bp.$8$\\ $\bigtriangleup$Bp.$2$\end{tabular}} & \textit{\begin{tabular}[c]{@{}l@{}}$\Diamond$St.$9$\\ $\Box$St.$5$\\ $\bigtriangleup$Bp.$2$\end{tabular}} \\ \hline\hline \begin{tabular}[c]{@{}l@{}}D. SVDD\\\cite{ruff2019deep}\end{tabular}
&\num{80.9}$\pm$\num{6}
&\num{94.29791409527297}$\pm$\num{2}& \num{87.40676469171618}$\pm$\num{5}& \num{86.6580472075934}$\pm$\num{3} & \num{93.88828824752821}$\pm$\num{3} & \num{85.94160849789546}$\pm$\num{6} & \num{82.8204190413711}$\pm$\num{2}  \\
\rowcolor[HTML]{EFEFEF}\begin{tabular}[c]{@{}l@{}}D. SVDD\\ w/ bias\end{tabular}
&\num{80.8}$\pm$\num{6}
&\num{94.12285563378088}$\pm$\num{2}& \num{85.37858505659713}$\pm$\num{5}&\num{82.1779408252072}$\pm$\num{3}  & \textbf{\num{94.58580337942063}$\pm$\num{2}}& \num{85.42917963298887}$\pm$\num{5} & \num{87.6253986746427}$\pm$\num{5} \\
\begin{tabular}[c]{@{}l@{}}D. SVDD\\ w/ bias \&\\noise reg.\end{tabular} 
&\textbf{\num{81.2}$\pm$\num{6}}
&\textbf{\num{96.34554485258727}$\pm$\num{1}}& \textbf{\num{87.86601628632157}$\pm$\num{4}}&\textbf{\num{87.943484111613}$\pm$\num{3}}  & \num{93.24102980286477}$\pm$\num{1} & \textbf{\num{90.75791262595772}$\pm$\num{4}}& \num{85.4348497980293}$\pm$\num{5} \\
\rowcolor[HTML]{EFEFEF} 
\begin{tabular}[c]{@{}l@{}}D. SVDD\\ w/ bias \&\\$\sigma^2$ reg.\end{tabular}
&\num{80.0}$\pm$\num{6}
&\num{89.38953898890862}$\pm$\num{4}& \num{81.11492954313044}$\pm$\num{6}&\num{83.57658185149519}$\pm$\num{4}  &\num{91.91700441550104}$\pm$\num{4} & \num{87.81709108543845}$\pm$\num{4}&\textbf{\num{88.6069412298648}$\pm$\num{4}} \\
\hline
\multicolumn{8}{c}{\textit{Bp. = Block perm.}\ \ \textit{St. = Strokes} ;\ \ $\Diamond$train set\ \ $\Box$val. set\ \ $\bigtriangleup$test set}
\end{tabular}
\end{center}
\end{table}

Lastly, we also show in Table \ref{Semi-supervised_one-class_tab}, the performance of the deep SVDD model and its variants with semi-supervised learning. 
For the regularized variants, we apply regularization to all labeled and unlabeled samples in a similar manner as the unsupervised regularized variants. 
For the one-vs-all setup conducted with the CIFAR10 dataset, the \textit{D. SVDD w/ bias \& noise reg.} model has the best performance. For the image alteration anomalies setup conducted only with the VOC2012 and WikiArt datasets, we ensure that the models are not trained and tested with the same type of anomalies to study the generalization capability of the models to a different type of anomaly. In the first two columns for VOC2012 and WikiArt datasets, the models were trained and validated/tested with different type of anomalies (\textit{block permutation} vs. \textit{strokes}). It can be seen that the models trained with \textit{block permutation} train set show better performance on \textit{strokes} test set, as compared to models trained with \textit{strokes} train set and tested on \textit{block permutation} test sets. 
Similar to unsupervised learning, the semi-supervised noise-regularized variant generally achieves better performance than the variance-regularized variant. For almost all settings, it can be observed that the proposed regularizers (mostly \textit{D. SVDD w/ bias \& noise reg.}) perform notably better than the other deep SVDD models.



\section{Conclusion}\label{CONCLUSION}
In this work, 
we considered two simple and effective regularization methods against mode collapse in deep SVDD. Even though the minibatch variance regularization method provided more control over the amount of regularization applied, the regularization method based on random noise injection generally performed better. Additionally, we demonstrated that both regularization methods performed better than the state-of-the-art GEO method on the \textit{strokes} test sets that do not portray apparent geometrical structures. We also showed that either one of the regularized variants achieved better performance than the standard deep SVDD and other classification-based models in a majority of the experiments. Moreover, we learned that the standard deep SVDD model and its variants performed better and were more stable at low learning rate. \changemarker{As noted by one of the reviewers, our proposed regularizers could be of interest for mitigating the mode collapse problem in GANs.}

\changemarker{\section*{Acknowledgment}
PC acknowledges support by the ST Engineering Electronics-SUTD Cyber Security Laboratory. AB acknowledges support by the ST Engineering Electronics-SUTD Cyber Security Laboratory and Ministry of Education of Singapore (MoE) Tier 2 grant MOE2016-T2-2-154.
LR acknowledges support by the German Ministry of Education and Research (BMBF) in the project
ALICE III (01IS18049B). MK acknowledges support by the German Research Foundation (DFG) award KL 
2698/2-1 and by the Federal Ministry of Science and Education (BMBF) 
awards 031L0023A, 01IS18051A, and 031B0770E.
}



\bibliographystyle{IEEEtran}
\bibliography{ijcnn2020_mode_collapse}


\end{document}